\documentclass[conference]{IEEEtran}
\IEEEoverridecommandlockouts

\usepackage{datetime}
\usepackage{csquotes}
\usepackage{tabularx}
\usepackage[table]{xcolor}
\usepackage{amsmath,amssymb,amsfonts}
\usepackage{algorithmic}
\usepackage{graphicx}
\usepackage{textcomp}
\usepackage{xcolor}
\usepackage[utf8]{inputenc}
\usepackage{caption}
\usepackage{multicol}
\usepackage{lipsum}
\renewcommand\thesection{\arabic{section}}
\usepackage{fancyhdr}
\usepackage[section]{placeins}

\usepackage[style=bwl-FU ]{biblatex}
\usepackage[utf8]{inputenc}
\usepackage[T1]{fontenc}
\usepackage{textcomp}
\usepackage{hyperref}
\DeclareUnicodeCharacter{2060}{\nolinebreak}
\usepackage{xspace} 
\usepackage{array}
\def\BibTeX{{\rm B\kern-.05em{\sc i\kern-.025em b}\kern-.08em
    T\kern-.1667em\lower.7ex\hbox{E}\kern-.125emX}}
    
\renewenvironment{abstract}
 {\small
  \begin{center}
  \bfseries \abstractname\vspace{-.7em}\vspace{0pt}
  \end{center}
  \list{}{
    \setlength{\leftmargin}{.7cm}%
    \setlength{\rightmargin}{\leftmargin}%
  }%
  \item\relax}
 {\endlist}
 
\begin{document}
\title{\textbf{\Large Window Detection In Facade Imagery:\\ 
A Deep Learning Approach Using Mask R-CNN}}
\author{Nils Nordmark, University of Gothenburg\\Mola Ayenew, University of Gothenburg}
\maketitle

\begin{abstract}
The parsing of windows in building facades is a long-desired but challenging task in computer vision. It is crucial to urban analysis, semantic reconstruction, life cycle analysis, digital twins and scene parsing amongst other building-related tasks that require high-quality semantic data. In this article, we investigate the usage of the Mask R-CNN framework to be used for window detection of facade imagery input. We utilize transfer learning to train our proposed method on COCO weights with our own collected dataset of street view images of facades to produce instance segmentations of our new window class. Experimental results show that our suggested approach can with a relatively small dataset train the network only with transfer learning and augmentation achieve results on par with prior state-of-the-art window detection approaches, even without post-optimization techniques. 
\end{abstract}

\section{\raggedright \textbf{Introduction}}
Gathering semantic data regarding building features from images is desired as manual gathering can often be a challenging, time consuming and resource intensive task. For example, information regarding location, shapes and sizes, materials of building structures such as windows, doors and balconies are often not available, can change over time and the data gathering is usually a manual process. 

Many different approaches have been applied to solve the gathering of semantic data within many different domains. Within building facade parsing this task has mostly been treated as a process of semantic segmentation of facade imagery into architectural structural classes such as windows, doors, balconies and so on. This semantic segmentation approach provides information about facade classes as it assigns each pixel with a class label. This means that it does not differentiate instances of the same object as each pixel is classified to belong to a particular class. Specifically for parsing of windows instances in building facades, we suggest an approach that can decouple segmentation and classification to detect windows in an image with a bounding box and segmentation mask for each window instance. 

With instance segmentation, we can analyze the windows independently with detailed and comprehensive information for each pixel and specifically to which window it belongs to. To separate the pixels of the same category into different instances, we propose an approach based on the Mask R-CNN framework.

\section{\raggedright \textbf {Related Work}}
Facade parsing has been studied for a long time and there exists a large body of work on how to deal with such a problem.

\subsection*{ \textbf{Earlier approaches}}
\cite{zhao2010rectilinear} proposed parsing ground-level images into architectural units which could be used for city modelling. \cite{wendel2010unsupervised} used Scale–Invariant Feature Transform (SIFT) and \cite{recky2010windows} used K-means clustering to tackle the same problem. 

Many others proposed a procedural shape grammar approach. This approach could at a high-level be explained as a hand-crafted set of rules of basic shapes which represents structured geometries, i.e. the structure of the object is encoded as a set of parametric rules. Models could then be generated by iteratively applying the rules on a starting shape. 

\cite{muller2007image} used a procedural modelling pipeline of shape grammars combined with image analysis to derive hierarchical facade subdivisions. \cite{han2008bottom} extended the work with an attribute grammar which used a more advanced bottom-up and top-down mechanism for recognizing objects. Their method was based on previous object recognition work, such as \cite{zhu2000integrating} and \cite{tu2005image}. 

\cite{teboul2010segmentation} combined machine learning algorithms such as Reinforcement Learning (RL) with procedural modelling and shape grammar. Although their method could be said to improve the works of \cite{muller2007image} and \cite{han2008bottom}, the learning process to some extent still was based on hand-crafted rules. 

\cite{mathias2011automatic} took another approach based on architectural styles for automatic classification. They tried to extend \cite{muller2007image} work with appropriate style information as they argued that it could be used to select a more appropriate procedural grammar for the task of building reconstruction. Their approach was sort of an ‘inverse’ procedural modelling, i.e. an attempt to not rely on the assumption that the building style was known and needed to be conducted as a manual task. Instead, they tried to automate the classification of architectural styles, by using facade images to train their classifier to distinguish between three distinct architectural styles. For classifying those specific styles they were quite accurate but it also required a rather specific set of circumstances to succeed. 

As most of the earlier approaches used hand-crafted features and traditional classifiers a common limitation was that the required hand-crafted features often failed to represent the diversity of more complex objects, i.e. could not handle deviations from these set of user-defined rules well. They also suffered in accuracy when the input image was more complex, such as street-view images where the image was not perfectly rectified. Another common limitation was that they often suffered from heavy computational costs and slow running times (\cite{bala2016brief},  \cite{koch2018visual}).

\subsection*{ \textbf{Convolutional Neural Networks Approaches}}
More recently, convolutional neural networks (CNNs) based models have gained in popularity due to their good performances. With the success of AlexNet in 2012, many semantic segmentation approaches which used to apply traditional machine learning algorithms have instead been primarily based on CNNs \cite{bala2016brief}; \cite{parthasarathy2017brief}. 

Improved models quickly followed AlexNet such as Segnet (\cite{badrinarayanan20151segnet}) which was one of the first to tackle the problem of semantic segmentation of a whole image. 

Another major milestone came when \cite{long2015fully} showed that a fully convolutional network (FCN) trained end-to-end, pixels-to-pixels could exceed then state-of-the-art semantic image segmentation methods by retaining spatial information of an image. In more detail, they used three pre-trained base models (AlexNet, VGGNet and GoogleNet) and transferred them from classifiers by replacing fully connected layers with convolutional layers. Since then a semantic segmentation model trend based on FCNs followed with improving results (\cite{sultana2001evolution}). 

\cite{chen2017deeplab} use dilated convolution instead of plain convolution to set new state-of-art results with their proposed DeepLab system. 

Recently, more automated methods and entirely based on deep learning have achieved remarkable improvements in image analysis, such as \cite{penatti2015deep} and \cite{nogueira2017towards}. 

\subsection*{\textbf{Deep Learning Approaches}}
Deep learning and specifically deep convolutional neural networks have shown to outperform other image analysis methods for traditional vision tasks (classification, object detection, semantic segmentation and instance segmentation) in a lot of benchmarks and a variety of fields, e.g. medicine, surveillance, economics, sociology. They have been applied to a diversity of tasks such as face detection, speech recognition, autonomous vehicles and so on.

\subsection*{\textbf{Deep Learning Approaches Related to Facade Parsing}}
Naturally, deep learning was also applied to building-related image analysis problems as well. Recently there are some examples of work where deep learning has been applied for facade parsing quite specifically and rather successfully. 

\cite{schmitz2016convolutional} method for semantic interpretation of facade images used a convolutional network and a key insight was that even smaller datasets could train the network when transfer learning could be employed. For validation, they used the eTRIMS dataset to measure the overall accuracy, precision and recall. The F1 score had an accuracy of 0.82± 0.09 for the facade categories, e.g. windows 0.86 ± 0.05 0.67.

\cite{liu2017deepfacade} pushed the development of facade parsing even further with their method DeepFacade which combined both the learning capacity of deep convolutional neural networks with rules and a loss function based on symmetry found in building facades structures. They trained an FCN-8s network with their novel loss function to obtain experimental results on both the ECP dataset and the eTRIMS dataset that significantly outperformed previous state-of-the-art methods. Pixel accuracies on the eTRIMS dataset reached as high as ~0.91 for windows. 

\cite{fathalla2017deep} used appearance and layout cues combined with the VGG-model. Their method only had a single object of focus and overall accuracy was on par with other methods, although supplementary materials had to be asked for an in-depth understanding of the results. 

\cite{koch2018visual} developed a multi-scale patch-based pattern extraction that combined with a CNN (AlexNet), to automatically estimate the condition of buildings. They showed to some degree their method could serve as a proxy for condition estimates by appraisers. 

\cite{bacharidis20203d} used stereoscopic image and tachometer data combined with a deep learning-based facade segmentation stage based on generative adversarial networks (GANs). The task was to reconstruct 3D models of buildings facades, in the urban environment for cultural heritage. 

\cite{hu2020fast} used the bounding boxes detected by YOLO architecture in real-time to guide facade reconstruction in an interactive environment. Results showed accuracies above 0.82, which they deemed as sufficient for their 3D reconstruction task. 

\cite{rahman2020} developed a hybrid pipeline for his master thesis which consisted of both traditional and modern machine learning approaches. His system was trained for windows delineation segmentation, i.e. detecting window frames, mullions, transoms and so on. For small datasets it showed promising results, however, this method is not competitive compared to complete end-to-end state-of-the-art deep learning approaches such as \cite{liu2017deepfacade}. 

\cite{ma2020deep} used Faster R-CNN architecture for window detection, they specifically wanted to obtain the locations of the windows in their images for 3D reconstruction. They called their method WD-Net which showed ~0.5 per cent improved accuracy compared with the baseline Faster R-CNN. 

\section{\raggedright \textbf {OUR APPROACH}}
\subsection*{\raggedright\textbf{Mask R-CNN Architecture}}
In this paper, we suggested the Mask R-CNN model, which is an intuitive extension of the Faster R-CNN model and outputs the predicting segmentation masks on each Region of Interest (RoI), in parallel with the existing branch for classification and bounding box regression. Faster R-CNN is a popular framework for object detection but it is not designed for pixel-to-pixel alignment between network inputs and outputs. The segmentation mask in Mask R-CNN represents the pixel-level segmentation and alignment of each object that helps to extract each instance of an object separately without background. Mask R-CNN is a two-stage framework. The first stage scans the image and generates proposals, the second stage classifies the proposals and generates bounding boxes and masks \cite{mask1}; \cite{he2017mask}. 

\begin{figure}[hbt!]
\centering
\caption{\textit{Mask R-CNN framework}}
\includegraphics[width=7cm,height=4.8cm]{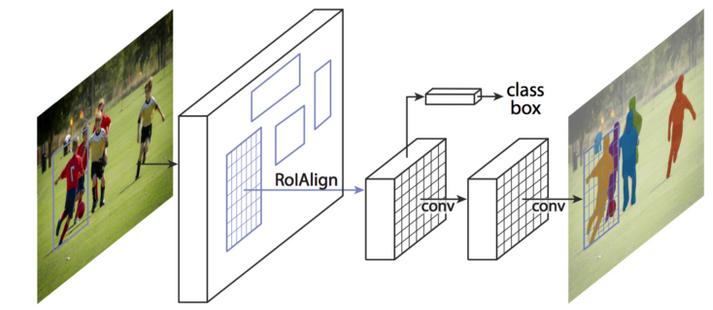}
\flushleft
\text{Note. Image from (\cite{he2017mask})}
\label{tab:ecp dataset}
\end{figure}

Mask R-CNN architecture consists of four high-level modules those are; Backbone, Region Proposal Network (RPN), ROI Classifier \& Bounding Box Regressor, and Segmentation Masks as shown in the (Figure \ref{tab:internalarchit}) below.

\begin{figure}[hbt!]
\centering
\caption{\textit{Mask R-CNN Internal Architecture}}
\includegraphics[width=7cm,height=7cm]{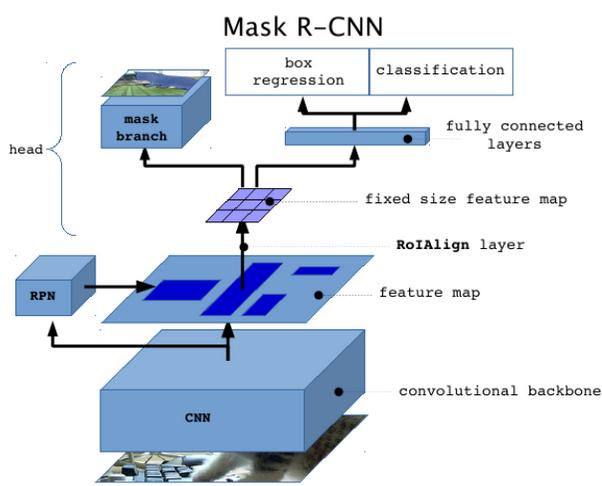}
\flushleft
\textit{Note. Figure 2 shows the steps until the instance segmentation processes takes place from the first stage to the last stage.}
\label{tab:internalarchit}
\end{figure}

\subsection*{\textbf{Backbone}}
The input image is fed into the CNN architecture, called the backbone. It is a pre-trained standard convolutional neural network, which is either ResNet50 or ResNet101. The backbone network serves as a feature extractor over an entire image, at the early layers the backbone detects low-level features like edges and corners, and later layers successively detect higher-level features like the car, person, sky. When the image passes through the backbone network, it is converted from 1024x1024px x 3 (RGB) to a feature map of shape 32x32x2048. This feature map becomes the input for the following stages.

Another more effective backbone, called a Feature Pyramid Network (FPN) applied on the top of the ResNet backbone to improve the feature extraction method. FPN uses a top-down architecture with lateral connections to build an in-network feature pyramid from a single-scale input. It improves the standard feature extraction pyramid by adding a second pyramid that takes the high-level features from the first pyramid and passes them down to lower layers. By doing so, it allows features at every level to have access to both, lower and higher-level features. Using a ResNet-FPN backbone for feature extraction with Mask R-CNN gives excellent gains in both accuracy and speed \cite{mask1}⁠.

\subsection*{  \textbf{Region Proposal Network (RPN)}}
A Region Proposal Network (RPN)  was introduced by \cite{ren2016faster} that takes an image as input and outputs a set of rectangular object proposals and finds areas that contain objects. To generate region proposals, the author scans the images in a sliding-window fashion over the convolutional feature map received from the ResNet101 + FPN backbone layers. The network takes as input an n × n spatial window of the input convolutional feature map. Each sliding window is mapped to a lower-dimensional feature. The RPN doesn’t scan over the image directly, instead, the RPN scans over the backbone feature map. This allows the RPN to reuse the extracted features efficiently and avoid duplicate calculations.Then this feature is fed into two fully connected layers, a box-regression layer, and a box-classification layer⁠.

\begin{figure}[hbt!]
\centering
\caption{\textit{Region Proposal Network (RPN)}}
\includegraphics[width=7cm,height=5cm]{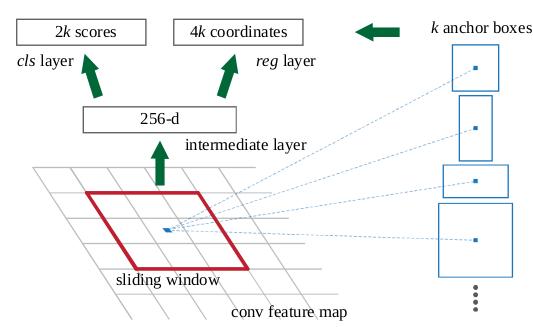}
\flushleft
\text{Note. Image from (\cite{ren2015faster} )}.
\textit{The RPN proposed by passing a sliding window over the CNN feature maps and at each window outputting k potential bounding boxes and scores for how good each bounding box expected to be.}
\label{tab:rpn}
\end{figure}

RPN has a classifier(cls) and a regressor(reg) as shown in the (Figure \ref{tab:rpn}). The regions that the RPN scans over are called anchors, which are boxes distributed over the image area. The anchor is the central point of the sliding window. For the ZF Model which was an extension of AlexNet, the dimensions are 256-d and for VGG-16, it was 512-d. The classifier determines the probability of a proposal having the target object. Regression regresses the coordinates of the proposals. For any image, scale and aspect-ratio are two important parameters. Aspect ratio is width of image/height of the image, the scale is the size of the image. (Figure 3) shows that if 3 scales and 3 aspect-ratio are chosen,  the total of 9 proposals are possible for each pixel, this is how the value of k is decided, K=9 for this case, k being the number of anchors. For the whole image, the number of anchors will be  W*H*K, where W and H are  the width and height of the image respectively \cite{karmarkar2018regional}.

The RPN generates two outputs for each anchor:\\
\textbf{Anchor Class: } The anchor class can be either foreground or background classes. The foreground class implies that there is likely an object in that box.

\textbf{Bounding Box Refinement: }A foreground anchor might not be centered perfectly over the object. So the RPN estimates a delta change to refine the anchor box to fit the object better. Using the RPN predictions, the model picks the top anchors that are likely to contain objects and refine their location and size.If several anchors overlap too much, it keeps the one with the highest foreground score and discards the rest (Non-max Suppression). After getting the final proposals (regions of interest) then, it passes to the next stage.

\begin{figure}[hbt!]
\centering
\caption{\textit{Example, a) Anchor boxes per object b) Top three anchor boxes pre object}}
\includegraphics[width=4.3cm]{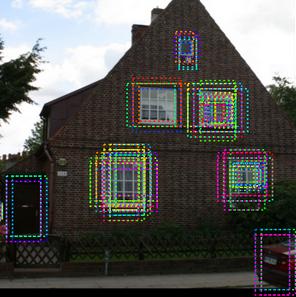}\hfill
\includegraphics[width=4.3cm]{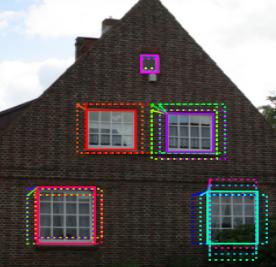}
\textit{(a)}\hspace{4cm}\textit{(b)}
\flushleft
\textit{ (a) shows that there are about 200k anchors can created for each object with different sizes and aspect ratios and they overlap each other to cover the object as much as possible. (b) shows that  the top three anchor boxes that are more likely to contain an object, after some refinement on  their location and size}
\label{tab:ecp dataset}
\end{figure}

\subsection*{\textbf{ROI Classifier \& Bounding Box Regressor}}
This  architecture was introduced by (\cite{girshick2015fast}), by removing the SVM classifiers for training to a regression layer and classification layer. This stage runs on the regions of interest (ROIs) proposed by the RPN.

First, the network processes all images with several convolutional and max-pooling layers to produce convolutional feature map. Then a region of interest (RoI) pooling layer extracts a fixed length feature vector for each object proposal from the feature map. Then, each feature vectored into the fully convolutional layers and produces softmax probability over different object classes with background class and output the probability of confidence for each object.

This module generates two outputs for each ROI:
\textbf{Class:} The class of the object in the ROI. Since this network is deeper and has the capacity to classify regions to specific classes like person, car, chair,etc. It can also generate a background class, which causes the ROI to be discarded.

\textbf{Bounding Box Refinement:} Further refine the location and size of the bounding box to encapsulate the object.

\textbf{RoI pooling layer:} Due to the bounding box refinement step in the RPN, the ROI boxes can have different sizes. ROI pooling is used to crop a part of a feature map and resize it to a fixed size. The RoI pooling layer uses max pooling to convert the features inside any valid region of interest into a small feature map with a fixed spatial extent of H×W (e.g., 7 × 7), where H and W are layer hyper-parameters that are independent of any particular RoI \cite{girshick2015fast}⁠. 

\subsection*{\textbf{Segmentation Masks}}
The mask network is the addition that the Mask R-CNN introduces. The mask branch is a convolutional network that takes the positive regions selected by the ROI classifier and generates masks for them. The generated masks are low resolution: 28x28 pixels. But they are soft masks, represented by float numbers, so they hold more details than binary masks. The small mask size helps to keep the mask branch light. During training, the ground-truth masks. scale down to 28x28 to compute the loss, and during inference, the predicted masks scale up to the size of the ROI bounding box and that gives us the final masks, one per object.

\subsection*{\textbf{Mask R-CNN Losses}}
The loss function in Mask R-CNN is the weighted sum of different losses at different stages of the model. Five general losses occur during training the weight \cite{mask1}.

\textbf{Rpn class loss:}  Occur when an improper classification of anchor boxes by the Region Proposal Network. This loss increases when the multiple objects are not being detected by the model in the last output. This loss determines how well the Region Proposal Network separates background with objects.\\
\textbf{Rpn bbox loss:} This loss occurs during the localization accuracy of the RPN. The object is being detected but the bounding box is not properly fit with the position of the object.\\
\textbf{Mrcnn class loss:} Occurs when an improper classification of objects that are present in the region proposal. This is to be increased in case the object is being detected from the image, but misclassified. It determines how well the Mask RCNN recognizes each class of object.\\
\textbf{Mrcnn bbox loss:} This loss occurs during the localization of the bounding box of the identified class. The loss increases when the correct classification of the object is done, but localization is not precise.\\
\textbf{Mrcnn mask loss:} This loss corresponds to masks created on the identified objects, which is related to how well the Mask RCNN segment objects. 

\subsection*{\textbf{Public datasets}}
As for current dataset resources, the Ecole Centrale Paris (ECP) Facades dataset \cite{teboul2010segmentation} and the eTRIMS \cite{korc2009etrims} database are the two most relevant public facade datasets we are aware of as both have been used as a benchmark for facade parsing in previous literature. 

The ECP dataset consists of 104 annotated images which are rectified and cropped facades of Haussmannian style buildings in Paris. Images are annotated with seven classes (balcony, chimney, door, roof, shop, wall and window). Ground truth object segmentation assigns each pixel to either one of the annotated objects or background. The object segmentation is represented as an image that consists of a colourmap (Fig. 5). The original annotations were improved by \cite{mathias2011automatic} to better fit the ground truth. 

\begin{figure}[hbt!]
\centering
\caption{\textit{Example from ECP dataset, a) the input image b) the annotation}}
\vspace{2mm}
\includegraphics[width=4.3cm]{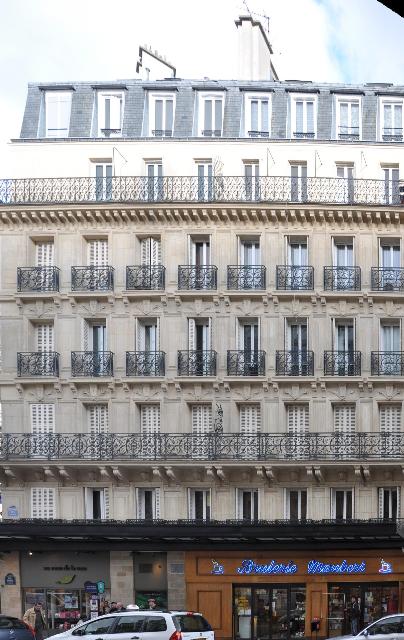}\hfill
\includegraphics[width=4.3cm]{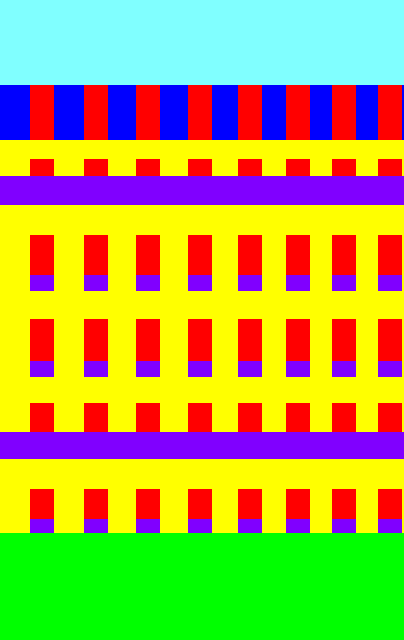}\\
\textit{(a)}\hspace{4cm}\textit{(b)}

\text{Note. Image from (\cite{teboul2010segmentation})}
\vspace{5mm} 
\caption{\textit{Example from eTRIMS, (a) the input image (b) the annotation}}
\vspace{2mm}
\includegraphics[width=4.3cm]{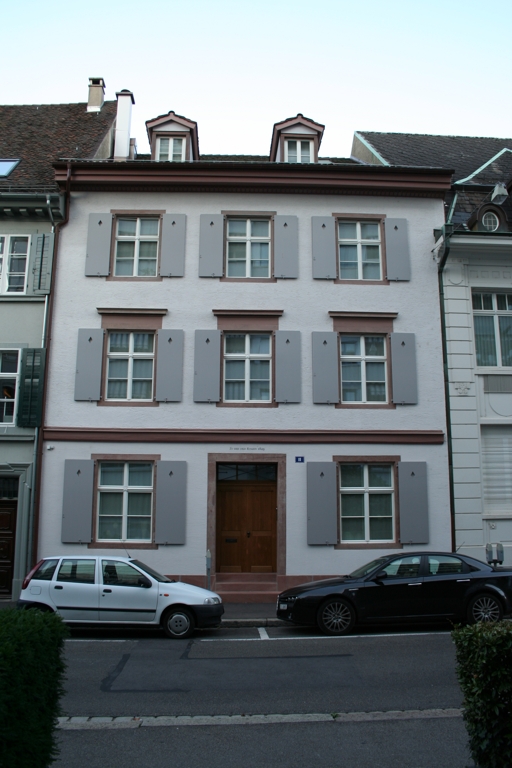}\hfill
\includegraphics[width=4.3cm]{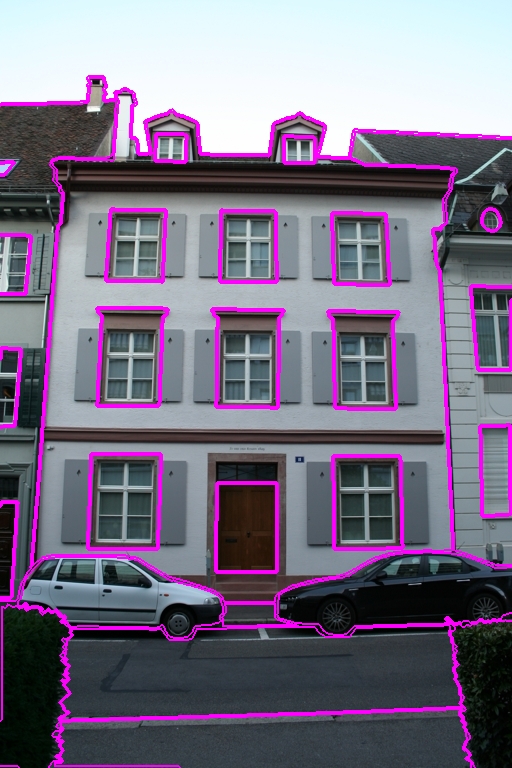}\\
\textit{(a)}\hspace{4cm}\textit{(b)}
\flushleft
\textit{Note. In the middle part of (b), the upper window boundaries include the ‘crown’ or flashing over the cornice. Image from (\cite{korc2009etrims}).}
\end{figure}

The eTRIMS consists of 60 annotated street view images but different from the ECP dataset the images are not rectified and the facades are of various architectural styles. Images are annotated with eight classes (sky, building, door, vegetation, car, road, pavement and window). Ground truth object segmentation assigns each pixel to either one of the annotated objects or background. The object segmentation is represented as an image that consists of an array and a color-map matrix (Fig. 6).

After inspecting these dataset we found it unwise to use either datasets for training our model to detect windows in real street view images because of multiple reasons. The ECP dataset only contains facades that are manually rectified and viewed in a front-parallel direction, thus using it as training data would not be wise since it is not highly representative of real street view images, which is the imagery we want to detect windows instances in. However, the most important reasons were that we found both to be imprecise or even wrongly annotated regarding what exactly defines the window boundaries. For example, the eTRIMS dataset often annotated the window boundaries to include not only the window frame but even other details such as the exterior casing (See Figure 6).

We searched and found a few more public datasets that contained window annotations but none were of very high quality and many had the same issues as mentioned above. Therefore we had to take a decision to build a quality dataset from scratch.

\subsection*{\textbf{Our dataset}}
We decided to use Google Street View images (Map data ©2020 Google) of facades mainly from central Gothenburg, Sweden. The cropped images were manually collected with an intention to contain a diversity of typical Swedish urban building facades that encompassed a large variety of window types. 

\begin{figure}[hbt!]
\centering
\caption{\textit{Two examples from images in our dataset annotated with the VIA tool
}}
\vspace{2mm}
\includegraphics[width=4.4cm,height=4.8cm]{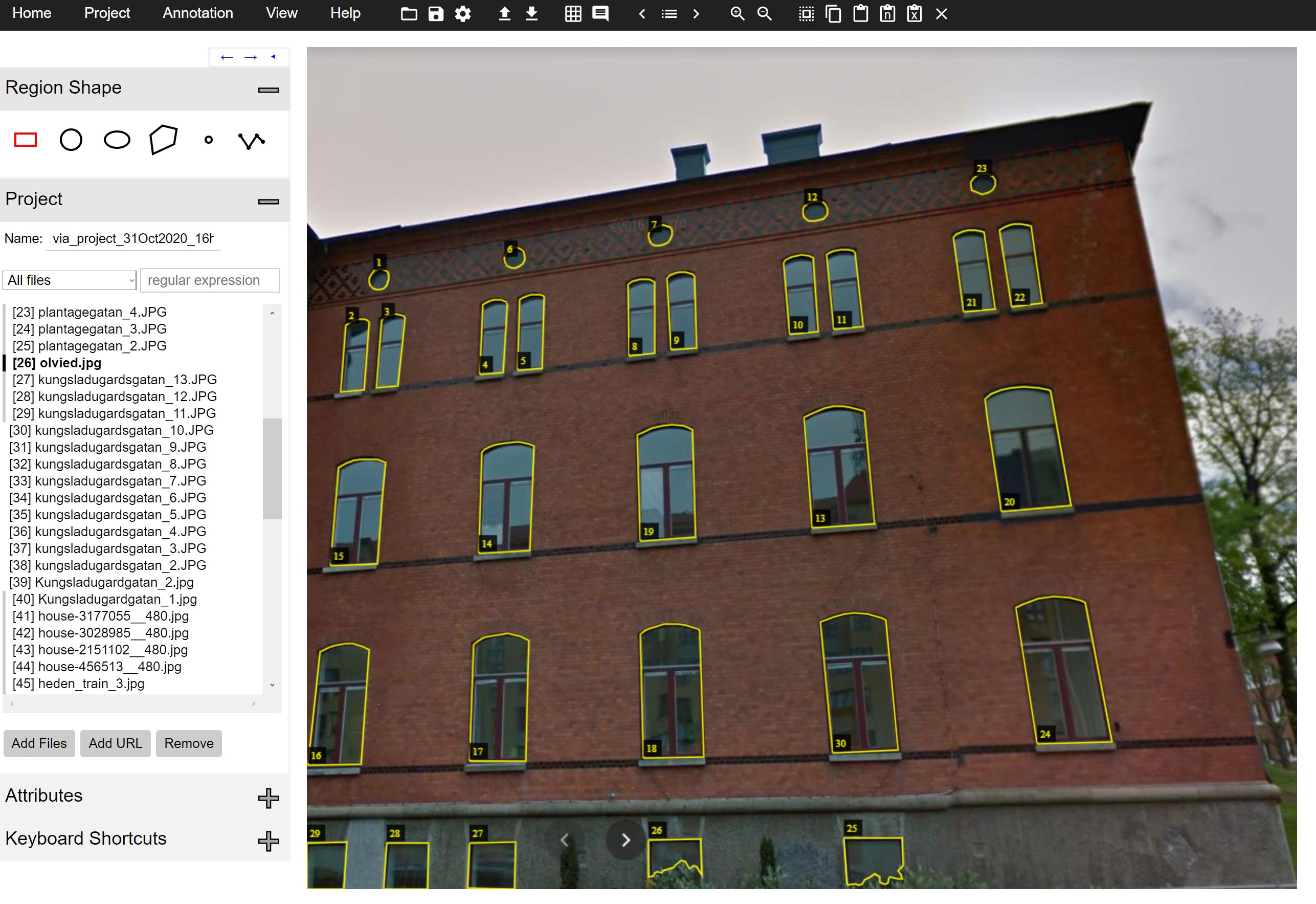}\hfill
\includegraphics[width=4.4cm,height=4.8cm]{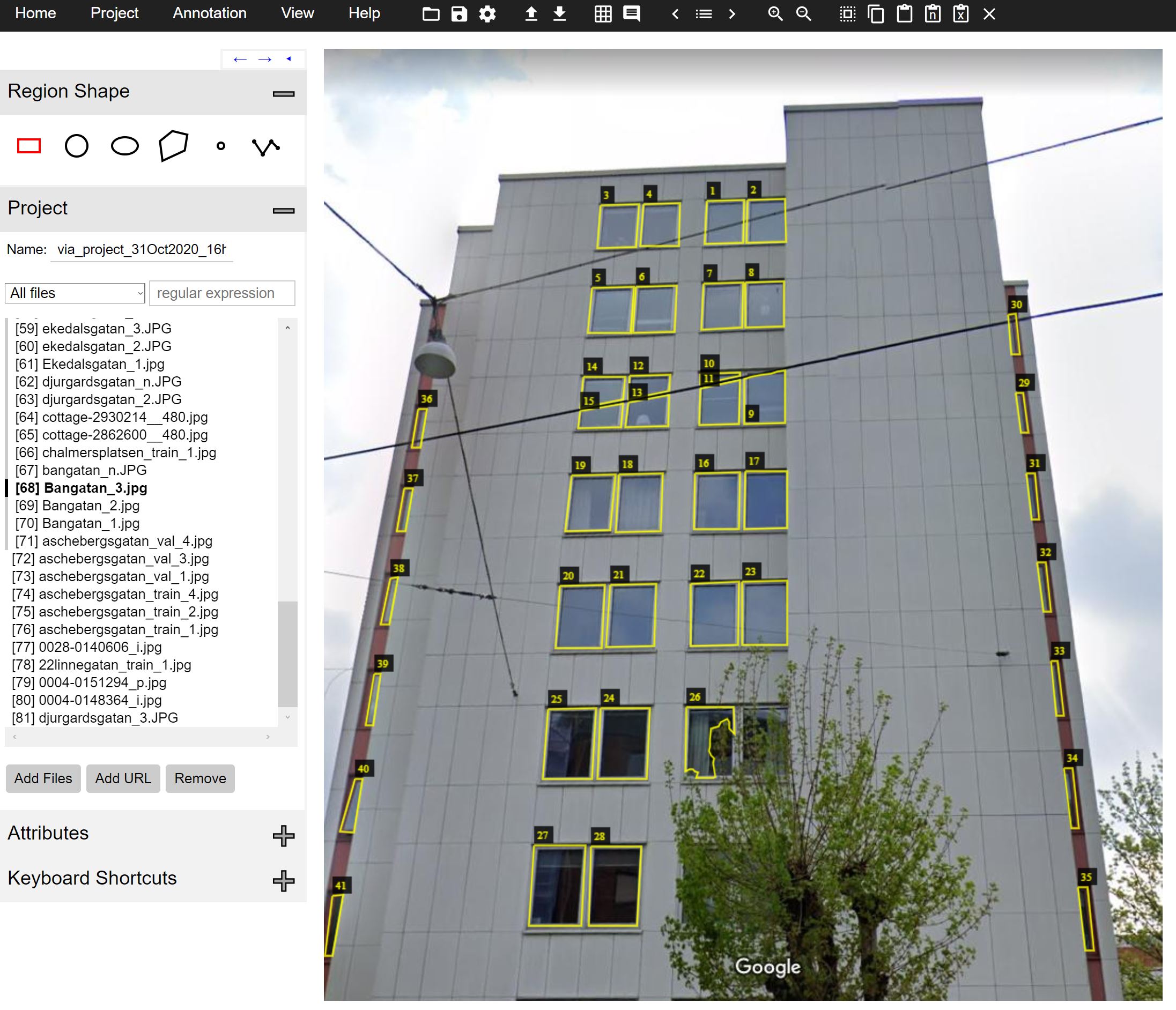}
\textit{(a)} \hspace{4cm} \textit{(b)}
\flushleft
\textit{Note. The yellow lines are the manually annotated polygons which define each window boundary.}
\label{tab:via}
\end{figure}

We annotated them manually with the VGG Image Annotator (VIA) tool \cite{dutta2019via}. There exists a lot of other tools to annotate images, e.g. LabelMe, RectLabel, LabelBox and COCO UI, but we chose VIA because of its efficiency and simplicity. For example, it runs in most modern web browsers so it does not require any other installation or setup. Regarding the annotated segmentation masks there is not a universally standard data format to store them. Some datasets save them as PNG images, while others store them in XML-format, and so on. The VIA tool saves its annotations in a JSON-format file where each mask is a set of polygon points (Figure \ref{tab:via}).

Since we tried to be consistent to follow each window's exact window frame and not to include other window parts as window boundaries it made annotation a very time-consuming and complex task, especially as specific decisions had to be taken for each window type. This precision and attention to detail made annotating the earlier images very gruelling, but once we got used to both the user interface and figured out most of the different window types and their frames, we were annotating one window about every 5-10 seconds. However, there were a lot of windows in some pictures and we wanted to be as consistent and precise as possible. 

\begin{figure}[hbt!]
\centering
\caption{\textit{Examples of the JSON-files generated with the VIA tool}}
\vspace{2mm}
\includegraphics[width=8cm,height=4.8cm]{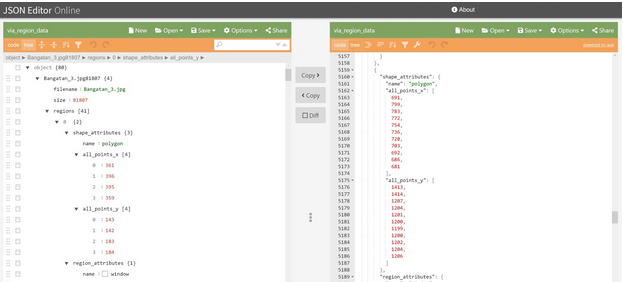}
\flushleft
\textit{Note. The left JSON-file corresponds to the training set and the right to the validation set. Both are viewed with JSON Editor Online.}
\label{tab:json}
\end{figure}

After gradually expanding our dataset we ended up with a collection of 100 images that contained 1540 annotated windows (an average of 15,4 windows per image). We then divide our dataset into a training set (80 images) and a validation set (20 images), i.e. we split the generated JSON-file into two different files (Figure \ref{tab:json}) with their corresponding images.

\subsection*{\textbf{Optimization Methods}}
To use the Mask R-CNN framework we used an open-source code implementation (Matterport) of it written on Python 3, Keras, and TensorFlow by \cite{abdulla2017mask}. Then we hosted the implementation on the Google Colab platform for free GPU usage and easy of shareability amongst other positive factors.

Assume to start with the structure of the Mask R-CNN as described in the paper \cite{he2017mask} and the source code provided in \cite{abdulla2017mask}, we needed to investigate the similarities and differences between the Matterport implementation running on Keras and TensorFlow and the official paper implementation build on the Caffe deep learning framework. 

From our understanding, the implementation followed the paper with only a few exceptions and those were done mostly for code simplicity and generalization. 

For example, to support training multiple images per batch, the Matterport implementation resizes all images to the same size and preserve the aspect ratio by padding with zeros if the image is not a perfect square. 

To support the training of different dataset’s annotations they chose to generate bounding boxes in the code on the fly, thus making image augmentations easier to apply. The downside is a slight decrease in bounding box placement accuracy,  about only ~2\% of bounding boxes differed by 1px or more. 

The paper uses a more aggressive learning rate of 0.02, but for the Matterport implementation that was too high and causes the weights to explode in TensorFlow as it computes gradients differently from Caffe. To mitigate this problem, gradient clipping was implemented.

With an understanding of the differences between Matterport’s and the paper’s features, parameters, hyper-parameters and so on, we could start to optimize our model. The optimization adjustments and modifications were updated continuously during the whole process, but we
can perhaps give some intuition on how they were done by describing them as iteratively separate steps. For example, optimization of the network structure with its corresponding configuration can be considered ’low level’, while optimizing training was done with more regards to the properties of our dataset. Finally, in inference optimization, we look at the outcome and try to adjust our result within the context of our research problem or goal. 
Furthermore, each step often needed distinct configuration adjustment depending on which model weights and dataset changes were used, i.e. training resumed with COCO-weights, augmentation flip, augmentation affine and so on, all needed different configurations to train properly.

\begin{figure}[hbt!]
\centering
\caption{\textit{Examples of the many configurations options in the Matterport implementation}}
\vspace{2mm}
\includegraphics[width=8cm,height=4.8cm]{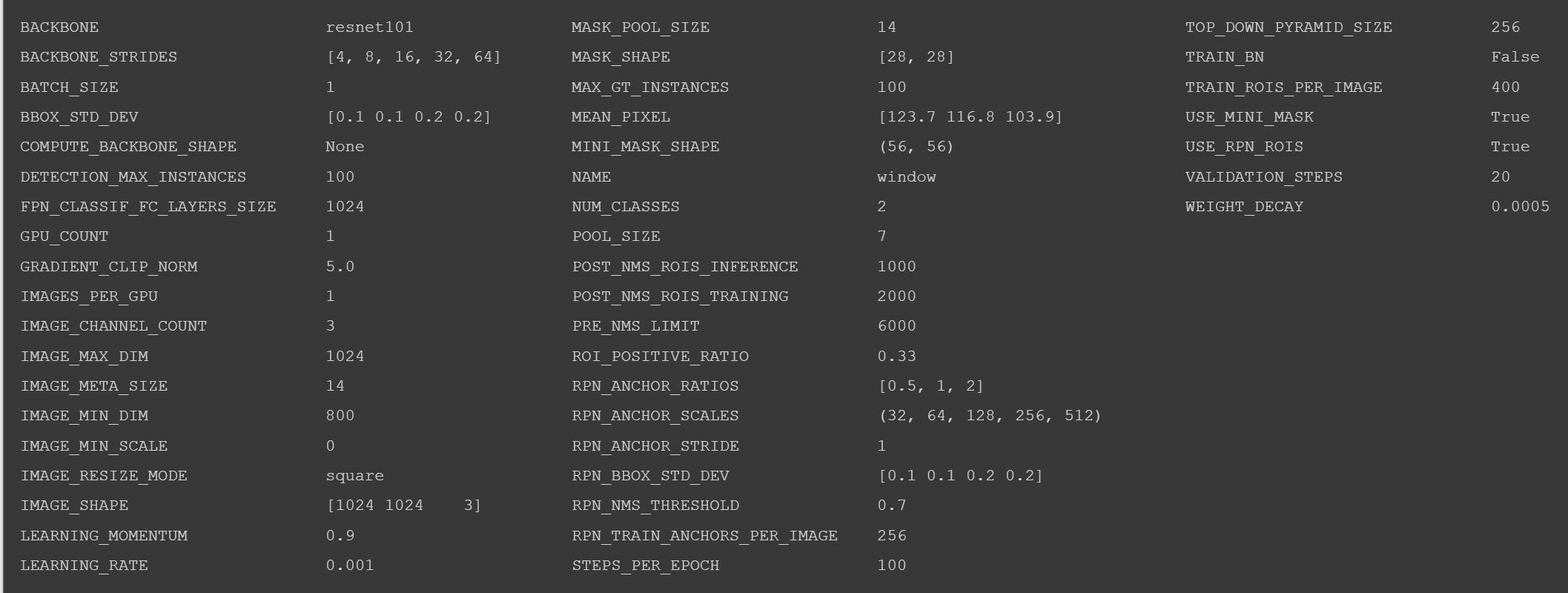}
\flushleft
\textit{Note. This configuration was used throughout the training steps for the model which achieved the highest mAP on our testset (See Table 1 and Figure 12).}
\label{tab:config}
\end{figure}

\section{\raggedright \textbf {EXPERIMENTS}}
In this section, we will first present some metrics used in our experiments, then give insight into the thought process of some of our structure and configuration optimization, training optimization and inference optimization. Finally, we will present our experimental settings and some of the best results we achieved.

\subsection*{\textbf{Metrics}}
To use the networks inference results for detecting windows and analysing the performance of our model, we are interested in detecting every existing window as well as reducing the number of wrongly detected windows. In other words, recall is a crucial value to improve. To evaluate the images overall detection results, we first use IoU (Intersection over Union) to measure how much our predicted boundary overlaps with the ground truth (the real object boundary). A predefined IoU threshold of 0.5 or greater is used for classifying the prediction as a true positive otherwise a false positive. 

\begin{figure}[hbt!]
\centering
\caption{\textit{Examples of IoU.}}
\vspace{2mm}
\includegraphics[width=8cm,height=4.8cm]{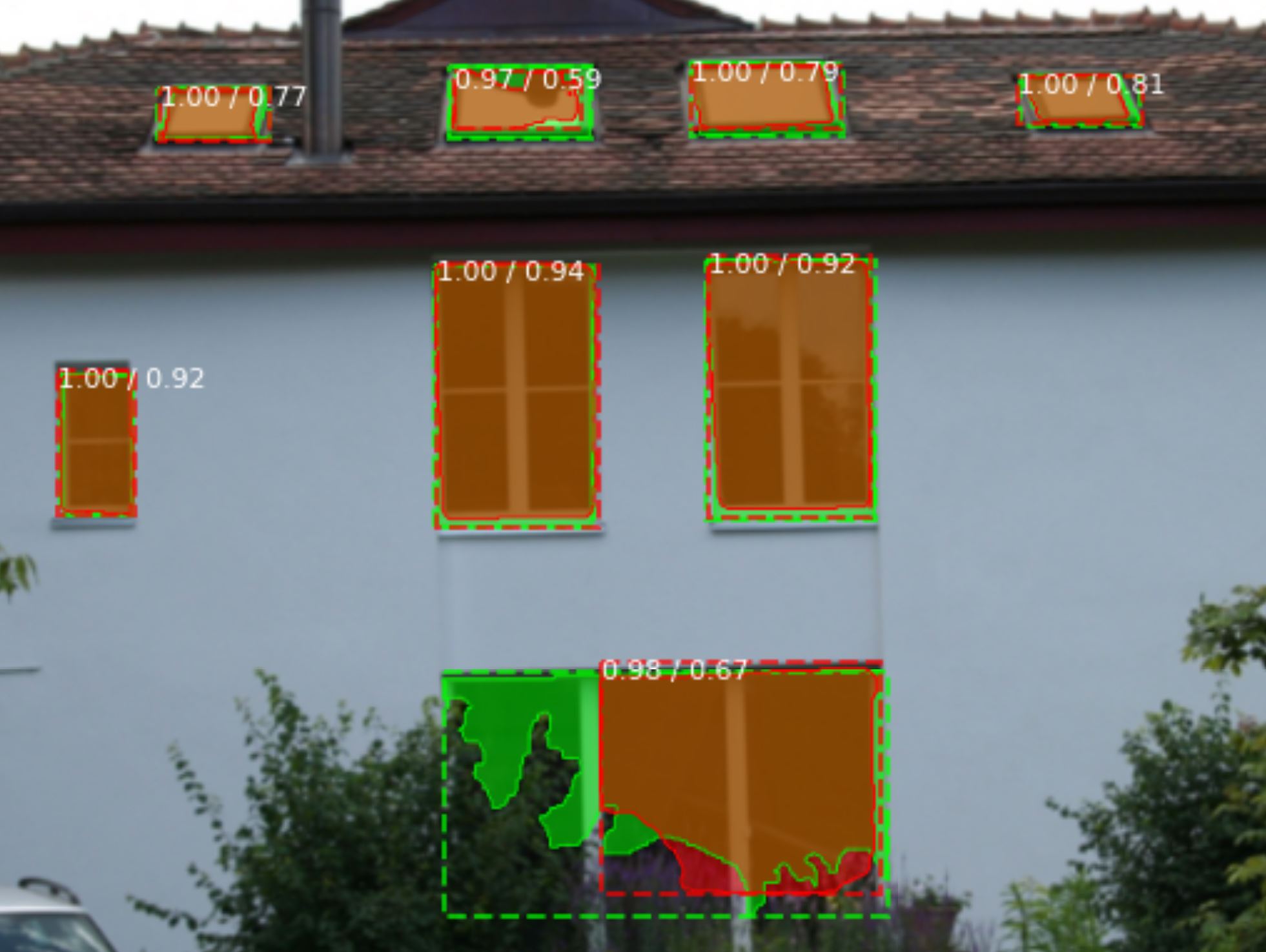}
\flushleft
\textit{Note. The ground truth is colored green and the prediction is colored red. Each detected window instance has a captions with its own confidence score and measured IoU.}
\label{tab:iou}
\end{figure}
Then we use the typical precision and recall rate:
\begin{equation} \label{eq1}
\begin{split}
Precision & = \frac{TP}{TP + FP} \\
\end{split}
\end{equation}
\vspace{2mm} 
\begin{equation} \label{eq1}
\begin{split}
Recall & = \frac{TP}{TP + FN}
\end{split}
\end{equation}
where, TP, FP, FN denotes the true positive, false positive and false negative, respectively.

Finally, we use the mean Average Precision (mAP):
\begin{equation} 
mAP I\& = \int_{0}^{1} P(R)\,dR 
\end{equation}
where the P represent the precision rate and and R the recall rate. 

\subsection*{\textbf{Structure and configuration optimization}}
Overall, we found it a challenging task to test and tune the many parameters, mainly due to the adaptive process of such a large search space, the continuous updates to our training data, but also the time it takes to train and do inference to calculate the accuracy. However, we will try to summarize the more important optimization decision we took.

We tested many different ‘lower-level’ configurations of the neural network, for example, we investigated Mask R-CNN in combination with both the ResNet50 and the ResNet101 Backbone as described in \cite{he2017mask}. Those experiments showed that the latter often yielded better results. 

We kept the base configuration of input images of size 1024x1024 px for best accuracy even if some of our images were a bit smaller since the model resized them automatically. Further testing could benefit separate tests with batches of different image sizes. 

We found the gradient clipping norm to be appropriately set a 5 with a learning rate at 0.001 and momentum of 0.9 and so on. 

By plotting each test we could observe that training and validation losses indicated overfitting, i.e. we got an increasingly lower training loss but the validation loss at certain points started to fluctuate or stagnate (Figure 11).

\begin{figure}[hbt!]
\centering
\caption{\textit{Examples of plots from our training}}
\vspace{2mm}
\includegraphics[width=9cm,height=4.8cm]{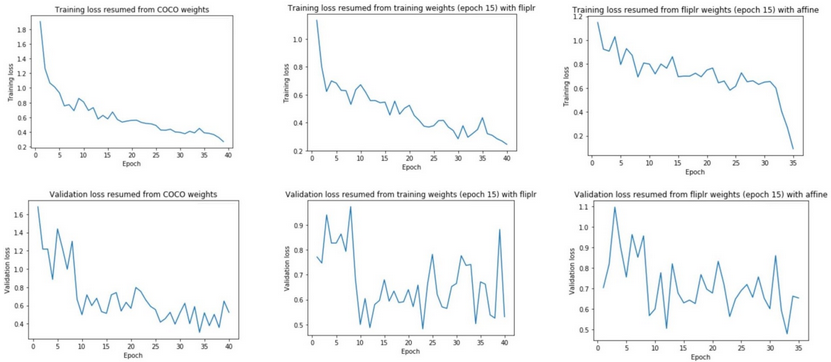}
\flushleft
\textit{Note. This particular plot was tested for one of our k-fold cross validation tests.}
\label{tab:losses}
\end{figure}

To overcome overfitting we experimented with weight decay and dropout amongst other things. Unfortunately, GridSearch with cross-validation, to automatically search for the optimal hyper-parameter values, required more computational power than we had access to, but we believe a valid option given the appropriate resources. 

Another option and the most intuitive, to prevent overfitting is often to extend the dataset. However, this was neither feasible given the remaining project time and also did not fit well with our interest in investigating the performance of a relatively small dataset.

Finally, we used k-fold cross-validation, which is a resampling procedure used to evaluate machine learning models on a limited data sample, without any major improvements on our test set in terms of accuracy.

\subsection*{\textbf{Training optimization}}
For the training optimization we assume the given Mask-RCNN loss function as:
\begin{equation}
L = \alpha L_{1} + \beta L_{2} +\gamma L_{3} + \delta L_{4} +  \epsilon L_{5}
\label{tab:equa4}
\end{equation}

where the first loss is the RPN class loss and the second the RPN bounding box prediction loss. The other three losses correspond to those of the head neural networks classes, bounding box and mask loss respectively. The training optimization goal is to find a corresponding configuration such that it minimizes the overall loss. This is an adaptive process of matching the results of the loss function with the given dataset and the parameters as a result of a trial process. 

Given a loss $L_i$ and equation (\ref{tab:equa4}), optimize $K = (\alpha,  \beta, \gamma, \delta, \epsilon) \in R^5$ and  $L_1,L_2,L_3,L_4, L_5$ loss functions so that the average precision of our model so that the average precision of our model is improved with a threshold of $AP_5{}_0$ :

\begin{equation} 
\max\limits_{K,L_{1},L_{2},L_{3},L_{4},L_{5}}
 AP_{50}
\end{equation} 
Ultimately, the final choice of parameters chosen depends on the structure configurations discussed in the previous section and also on our input data . While technically for K any combination of real values can be combined:
\begin{equation}
 (\alpha,  \beta, \gamma, \delta, \epsilon)\sim n.(\alpha,  \beta, \gamma, \delta, \epsilon),\forall n \in \mathcal{R}
\end{equation}%
This means that even though the total loss value changes, our accuracy of the detection is dependent on the loss gradient and not the loss value. Furthermore, several other configuration parameters need to be tested and changed to improve accuracy. The practical approach for tuning loss weights was to start with a default weight of 1 for each loss and evaluate the model on the validation set by inferring the model performance image by image. We had to evaluate the true positives, false positives, and false negatives combined with IoU of their masks and their score.

\begin{flushleft}
\textbf{Transfer learning}
\end{flushleft}

In our approach, we specify that training should start from weights pre-trained on COCO dataset. In other words, we are using transfer learning, which means we don’t need to train a new model from scratch and that instead utilize a lot of the already learned features from the COCO dataset, which contains around 120K images. After the training for 40 epochs with k-fold cross-validation have been completed, we save the model’s parameters and evaluate the new weights (one for each epoch is generated). 

\begin{flushleft}
\textbf{Augmentation}
\end{flushleft}

To train our model with more training data, we used augmentation on our existing data as annotating new images would take too long. Since the polygons are calculated internally this was easy to implement with the imgaug library.

First, we resume training, on the weights with the lowest training loss from the previous training, with augmentation that flip/mirror each image in the training set horizontally right/left. Then we again save the model’s parameters and evaluate the performance of the new weights. 

Then we resume training, on the weights with the lowest training loss from the previous training, with affine which transforms each image in the training set by rotation (-45, 45) and shearing (-16, 16). Finally, we save the model’s parameters and evaluate the performance of the new weights with running inference on the weights that yielded the lowest training loss. 

\subsection*{\textbf{Inference optimization}}
In the process of optimizing the trained network, we constantly adjusted and modified the training set and other parameters. Since the output of the inference is a list of possible detection boxes and masks, whose probabilities are larger than our set threshold of detection minimum confidence, we need to adjust accordingly with regards to the properties of the test set images and the overall accuracy. We found a detection minimum confidence of 0.9 to yield as good trade-off of the precision and recall values for our windows. 

\section{\raggedright \textbf{Results}}
We perform our test result experiments on the eTrims dataset and not on the ECP dataset. The reason being that the ECP dataset doesn’t contain street-view images and the images are all manually rectified and viewed in a frontal direction, i.e. making the window detection task relatively easy. 

However, to properly use the eTRIMS dataset with our model, we had to refine their annotation to overcome inaccuracies and mistakes, and also accommodate our defined window boundary which follows the window frame. We, therefore, annotated 10 randomly chosen images for our test set to evaluate our model’s performances. Therefore, any comparisons should note these differences.

Both the training and inference was done on Google Colab with the free Tesla K80 GPU of about 12GB. We have overall trained and evaluated more than 60 models with different configurations. Out of those we present the best three models with regards to the highest mAP on our test set (Table 1) and for visualization we present the inference done on the model achieving the highest mAP below (Figure 12).

\begin{table}[!htb]
    \caption{\\ \textit{A comparison of pixel accuracies on the eTRIMS dataset. Accuracies are shown in percentage.}}
    \vspace{2mm} 
    \begin{tabular}{m{1.3cm} m{0.4cm} m{0.4cm} m{0.4cm} m{0.4cm} m{0.4cm} m{0.7cm} m{0.7cm} m{0.7cm}}
    \textbf{Class}&\text{[1]}&\text{[2]}&\text{[3]}&\text{[4]} &\text{[5]} &\text{Our*} &\text{Our**} &\text{Our***}
    \\
    \hline
    \textbf{Window[\%]} & 75 & 73 & 71 & 86 & 90.91 & 91.93 & 94.55 & 94.76
    \\
    \hline
    \end{tabular} 
    \flushleft
    \vspace{4mm}
    \textit{Note: Our result is only from 10 randomly chosen images from the eTRIMS dataset with refined annotation.\\
    \vspace{2mm}
    * model trained on coco-weights then resumed with fliplr and affine using k-fold cross-validation (the resumed weights were chosen by the folds highest mAP)\\
    \vspace{2mm}
    ** model trained on coco-weights then resumed with fliplr using k-fold cross-validation (the resumed weights were chosen by the folds highest mAP)\\
    \vspace{2mm}
    *** model trained on coco-weights then resumed with fliplr (the resumed weights were chosen by the lowest training loss)
    }
 \label{tab:accuracy}
\end{table}

\begin{figure*}
    \caption{\\ \textit{Window detection results from a model trained on coco-weights, then resumed with fliplr.}}
    \vspace{2mm}\hspace{0.4mm}
    Image
    \hspace{32.3mm}
    Ground truth
    \hspace{22mm}
    Prediction
    \hspace{27mm}
    Overlap
    \\
    \phantom{image}
    \small
    \hspace{34.8mm}
    green colors
    \hspace{29.6mm}
    random colors
    \hspace{26mm}
    green/red colors
    \\
    \phantom{image}
    \small
    \hspace{37.6mm}
    captions: class
    \hspace{27mm}
    captions: class/score
    \hspace{18.2mm}
    captions: score/IoU
    \vspace{3mm}
    \\    
    \includegraphics[scale=0.142]{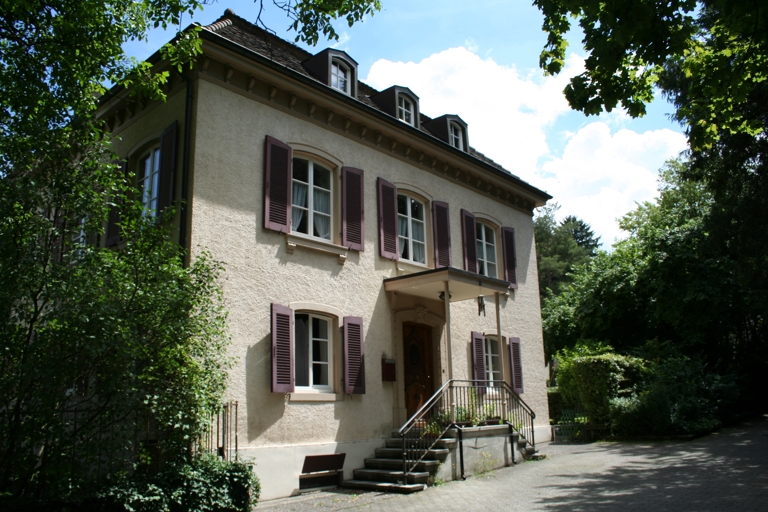}\hfill
    \includegraphics[scale=0.27]{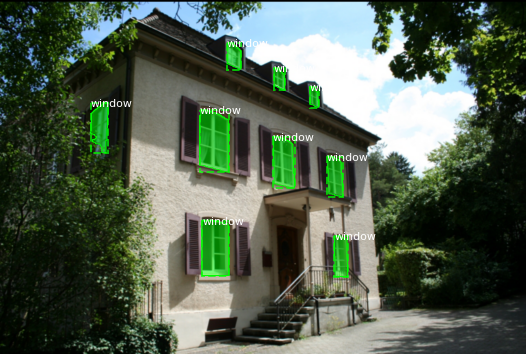}\hfill 
    \includegraphics[scale=0.27]{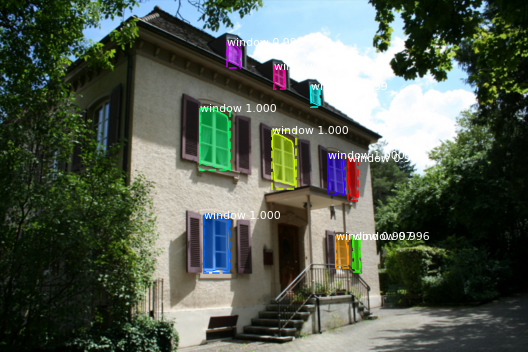}\hfill
    \includegraphics[scale=0.17]{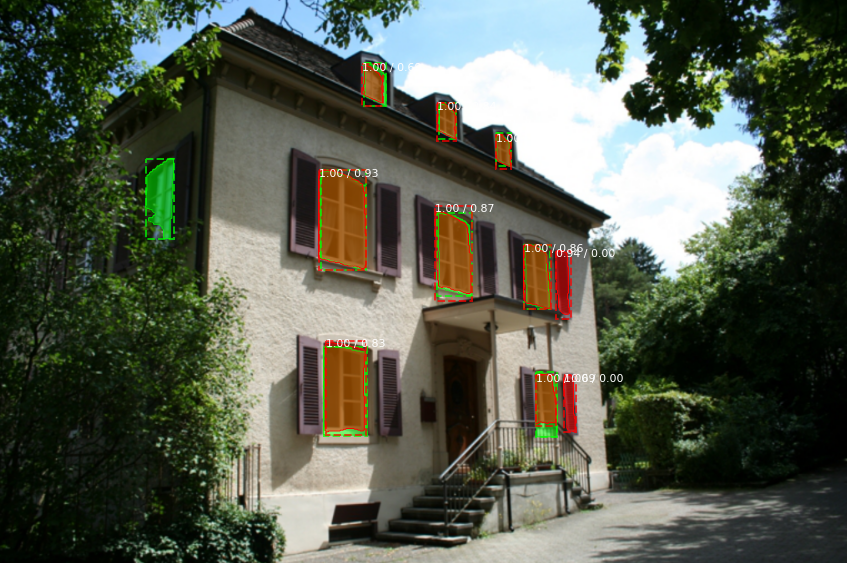} 
    \vspace{2mm} 
    \\
    \includegraphics[scale=0.137]{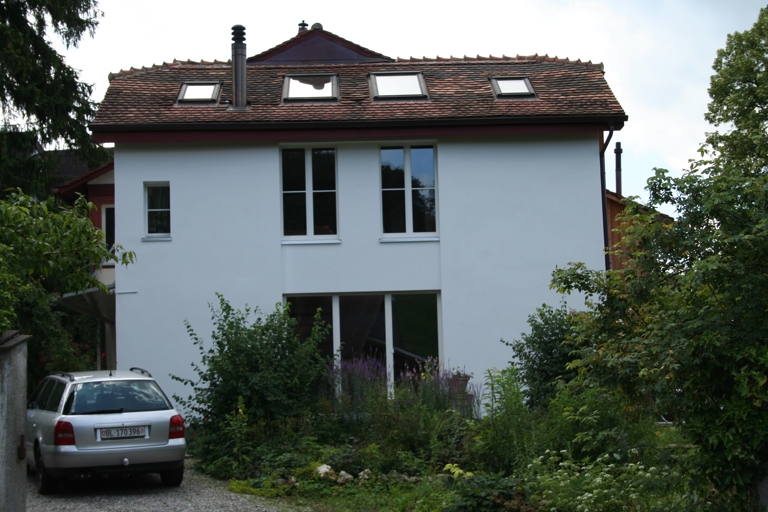}\hfill
    \includegraphics[scale=0.20]{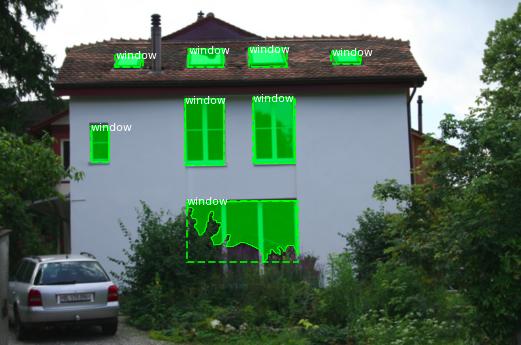}\hfill 
    \includegraphics[scale=0.20]{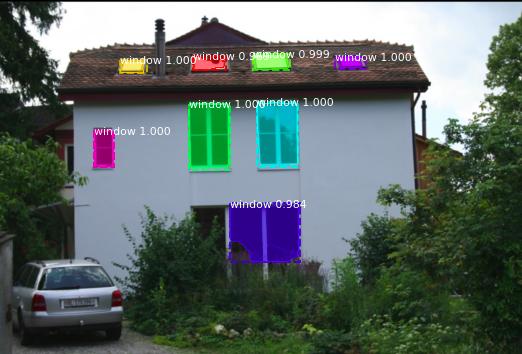}\hfill
    \includegraphics[scale=0.19]{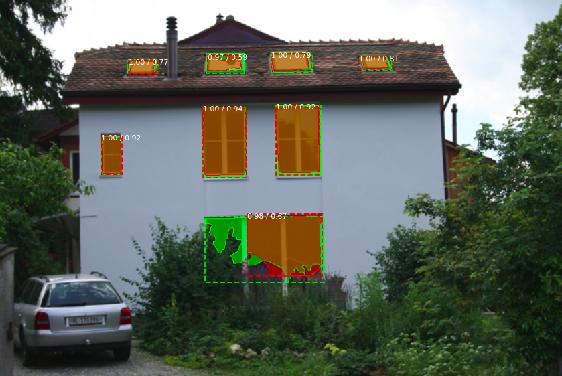} 
    \vspace{2mm}
    \\
    \includegraphics[scale=0.14]{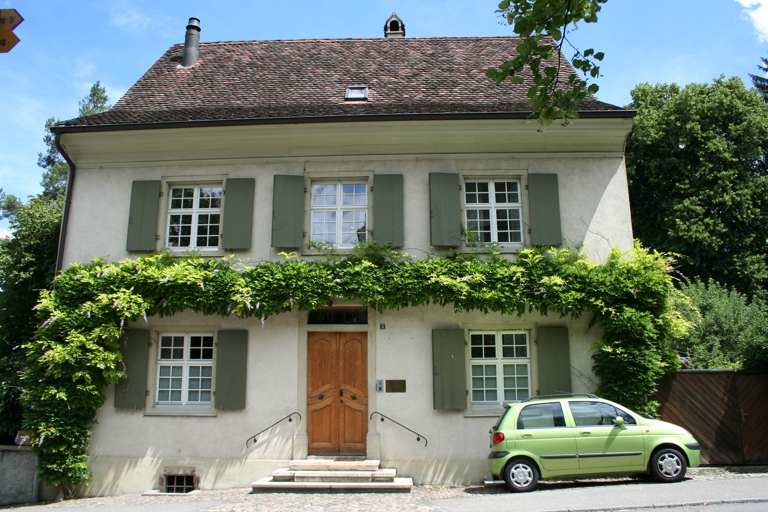}\hfill 
    \includegraphics[scale=0.27]{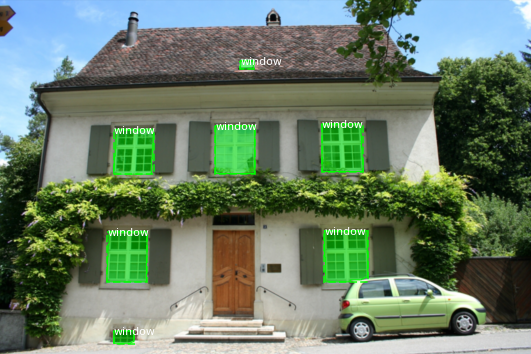} \hfill 
    \includegraphics[scale=0.27]{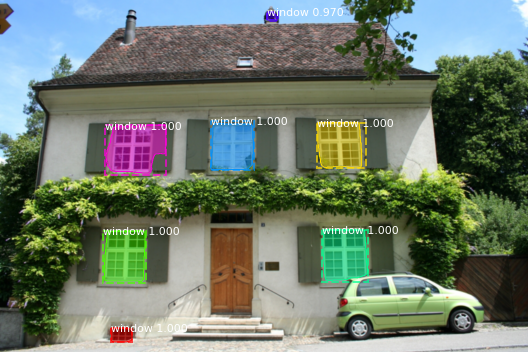}\hfill  \includegraphics[scale=0.165]{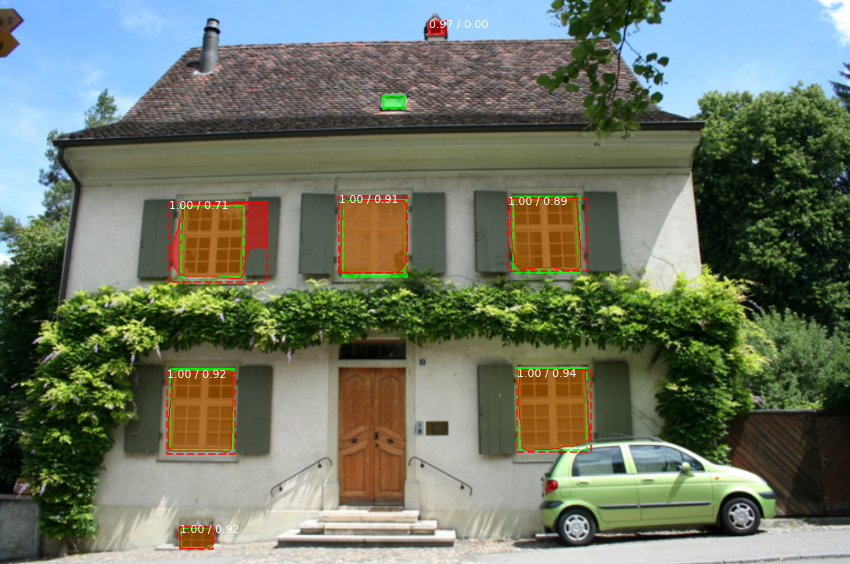} 
    \vspace{2mm}    
    \\
    \includegraphics[scale=0.12]{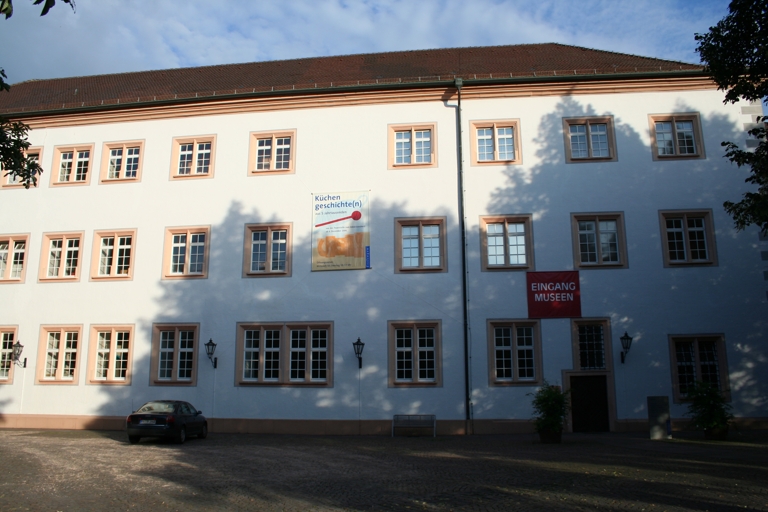}\hfill 
    \includegraphics[scale=0.235]{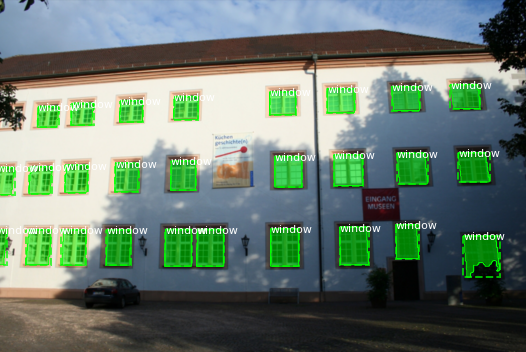} \hfill 
    \includegraphics[scale=0.235]{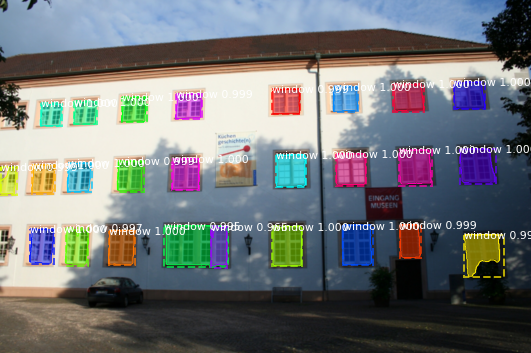}\hfill  \includegraphics[scale=0.15]{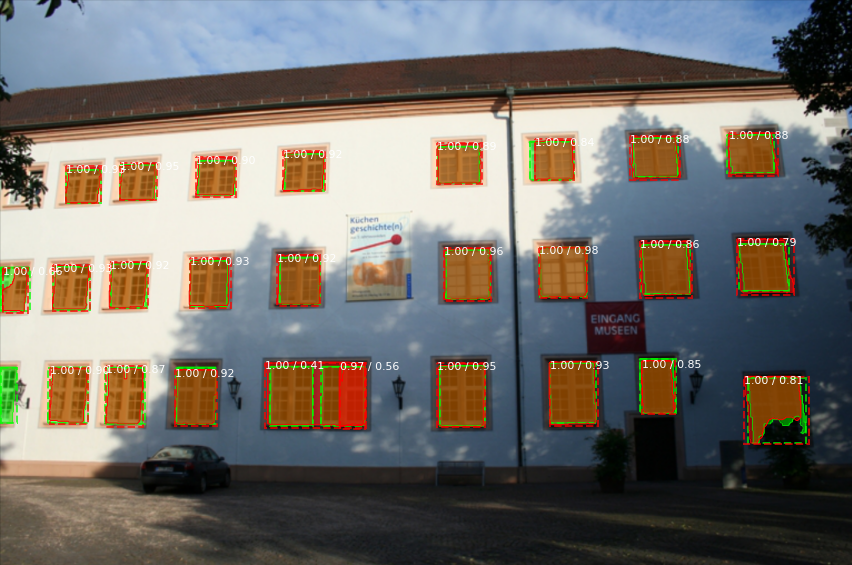} \\
    \vspace{2mm}
    \includegraphics[scale=0.12]{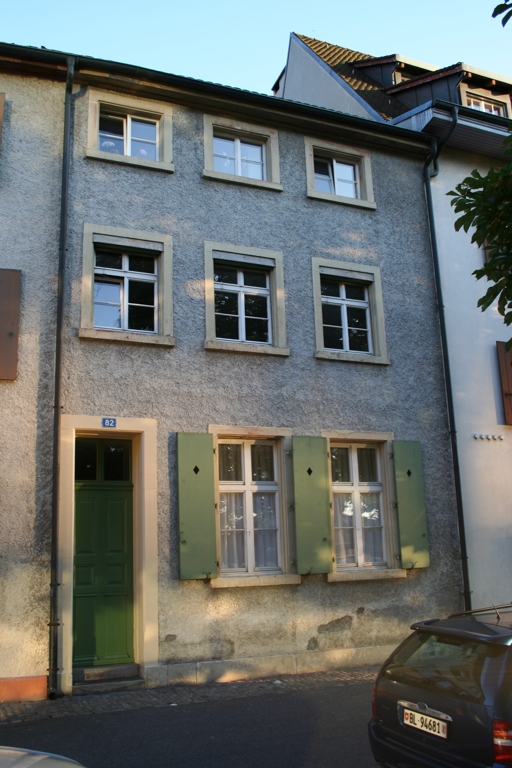}\hfill
    \includegraphics[scale=0.24]{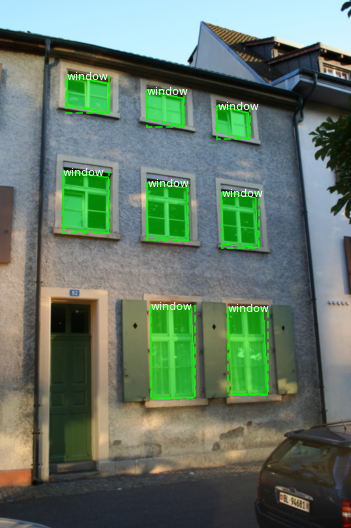}\hfill 
    \includegraphics[scale=0.24]{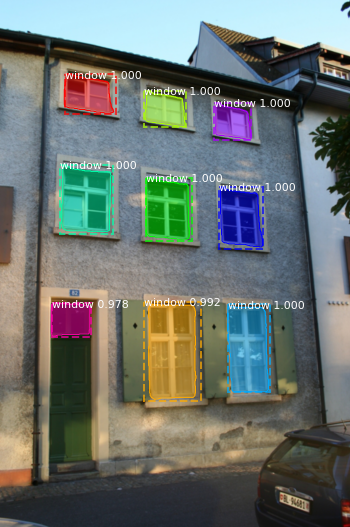}\hfill
    \includegraphics[scale=0.24]{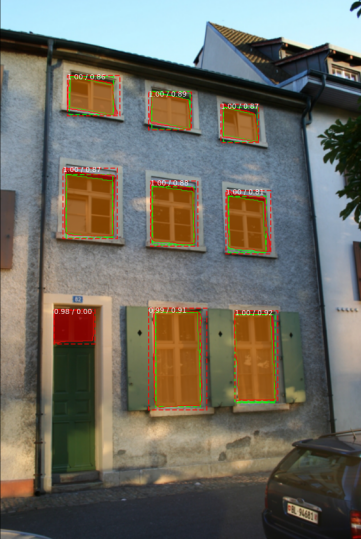} \\
    \vspace{2mm}
    \includegraphics[scale=0.10]{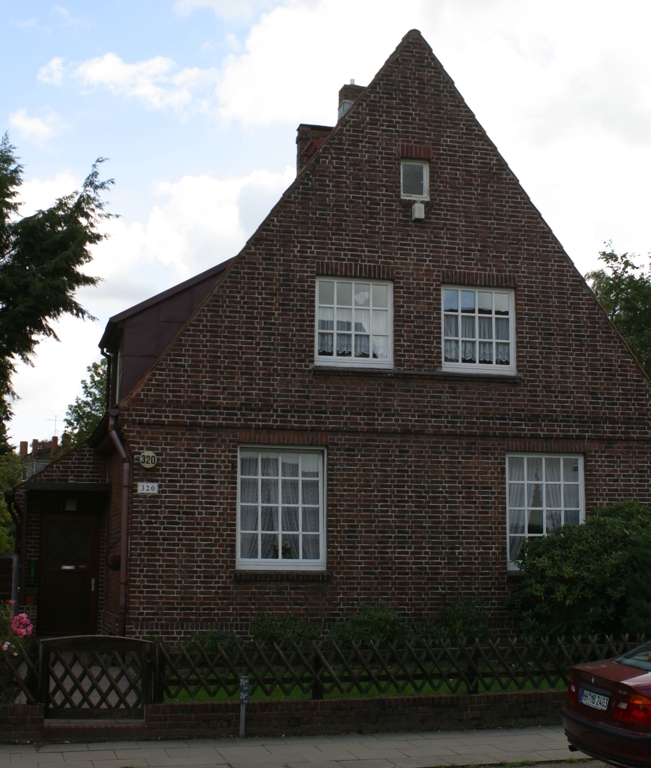}\hfill 
    \includegraphics[scale=0.21]{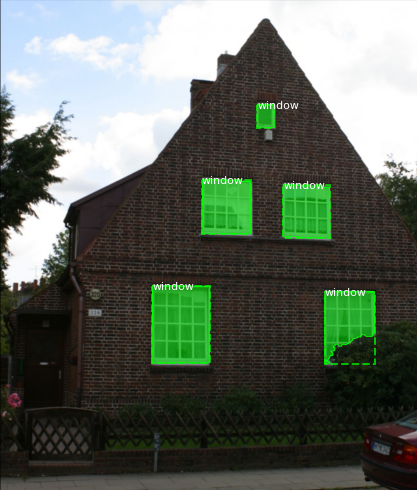} \hfill 
    \includegraphics[scale=0.21]{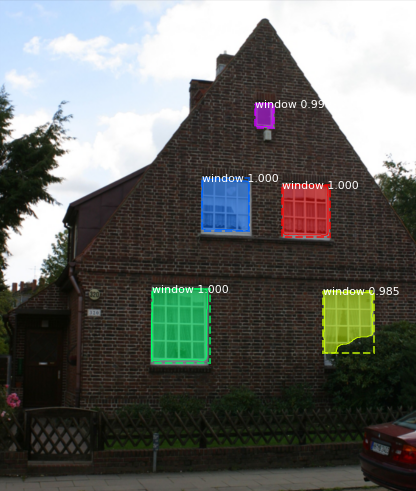}\hfill  \includegraphics[scale=0.13]{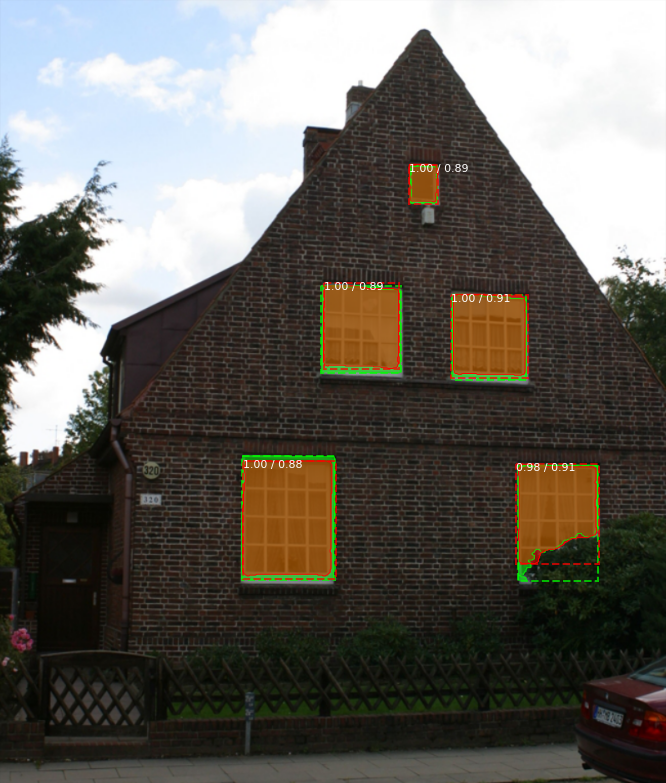} \\
    \vspace{2mm}       
    \includegraphics[scale=0.12]{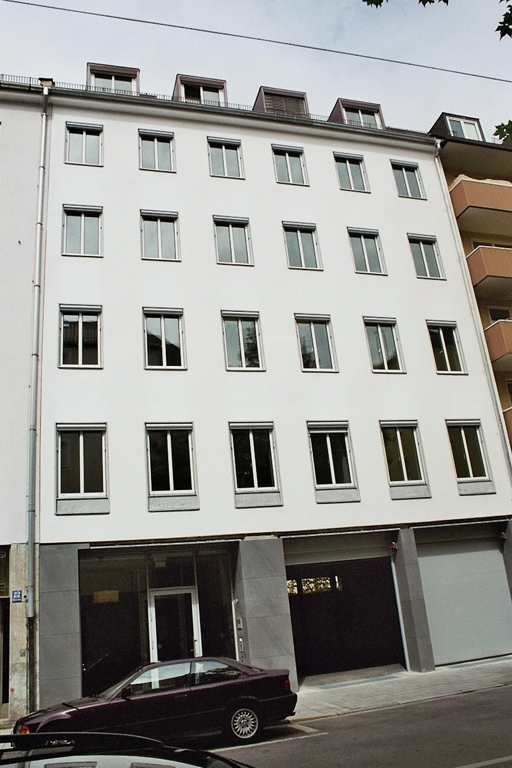}\hfill 
    \includegraphics[scale=0.23]{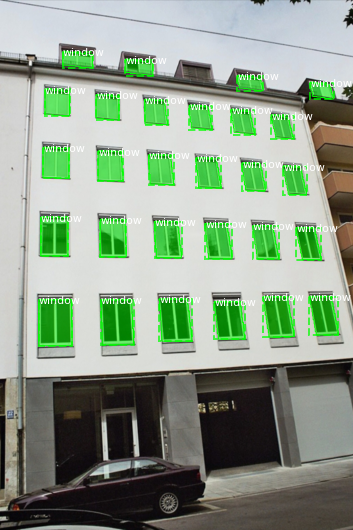} \hfill 
    \includegraphics[scale=0.23]{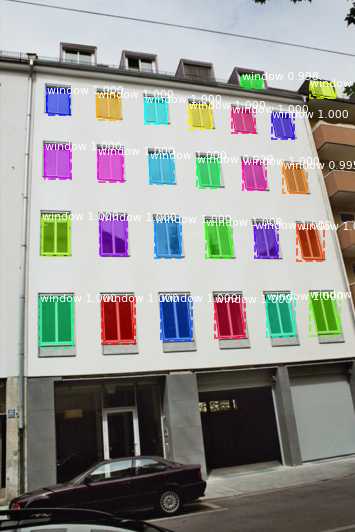}\hfill  \includegraphics[scale=0.15]{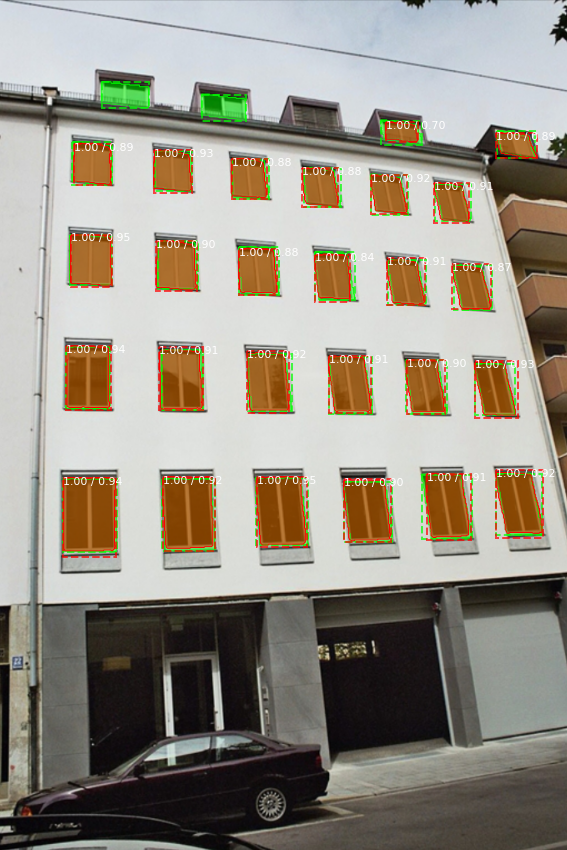} \\
    \vspace{2mm}

    \flushleft
\end{figure*}

\section{\raggedright \textbf {CONCLUSIONS}}
In this paper, we propose to use the state-of-the-art Mask R-CNN framework \cite{he2017mask} as an end-to-end network capable of instance segmentation without further machinery. Previous facade parsing detection work often only produced semantic segmentation of a categorical class (e.g. door, balcony, window etc.), while Mask R-CNN is capable of detecting and producing a bounding box and segmentation mask for each window instance. 

Experimental results show that our suggested approach can with only a relatively small but precise dataset train the network—only with transfer learning and augmentation—to produce instance segmentation of windows and a comparable result with prior state-of-the-art window detection approaches, even without pre- or post-optimization techniques.

As far as we know, we are the first to have annotated a street-view windows dataset by consistently trying to follow the window's frame as the window boundary and not to include other window parts. We believe that our consequent and precise annotation of windows helped our model training. We also hope it can benefit the research community as a free resource for more precise window detection. 

\section{\raggedright \textbf {FUTURE WORK}}
In this paper, we have got a sense of how to tune the hyper-parameters effectively with our training data. However, further work is recommended to investigate this with platforms, such as Wandb or Onepanel Jobs for easier coordination, synchronization and tracking of configurations. 

Another possibility is to extend the dataset with more high-quality annotations of window types were the model has low accuracy.

Further work could also benefit from training more facade classes, such as building, balcony, doors, chimney, to reduce false positives, false negatives. The same goes for the window class, i.e. splitting it into distinct window type classes, such as double-hung, transom etc. A model capable of a more complete facade parsing we believe will also make application easier as the model has a finer granularity of the facades properties. A caution is that this would require well-reasoned and well-defined class boundaries for each corner case of every facade structure. 

To further refine the results and perhaps mitigate some of the typical limitations we experienced, such as dealing with occlusion, overconfident false positives predictions, occasional mask overlaps, future work should investigate both in pre- and post-optimization methods of previous approaches or testing their novel approaches.  Also, code refactoring and optimization, e.g. rewriting some Python code in TensorFlow to bring all the benefits of the TensorFlow, Keras ecosystem to the model are desired. There is also always room for improvements.

\section*{\raggedright \textbf {ACKNOWLEDGMENTS}}
This paper was the final written report in the course DIT891 Project in Data Science. The course is offered within the Applied Data Science Master's Programme by the University of Gothenburg. We would like to thank Shirin Tavara, teacher and the course responsible and Richard Johansson, teacher. 

A special thanks goes to our supervisors Alexander Hollberg, Sanjay Somanath and Yinan Yu for their congeniality and support.

\printbibliography

\end{document}